\documentclass[letterpaper, 10 pt, conference]{ieeeconf}  % Comment this line out if you need a4paper

\IEEEoverridecommandlockouts                              % This command is only needed if
\usepackage{graphicx}
\usepackage{subfigure}
\usepackage{multirow}
\usepackage{amssymb}
\usepackage{algorithm}
\usepackage{algorithmic}
\usepackage{xcolor}
\graphicspath{
    {./fig/}
    }

\title{\LARGE \bf
A Configuration-Space Decomposition Scheme for Learning-based Collision Checking
}

\author{Yiheng Han$^{1}$, Wang Zhao$^{1}$, Jia Pan$^{2}$, Zipeng Ye$^{1}$, Ran Yi$^{1}$ and Yong-Jin Liu$^{1\dag}$% <-this % stops a space
\thanks{*This work was supported by the Natural Science Foundation of China (61725204)}% <-this % stops a space
\thanks{$^{1}$Y. Han, W. Zhao, Z. Ye, R. Yi and Y-J Liu are with BNRist, Department of Computer Science and Technology, Tsinghua University, Beijing, China.
        {\tt\small hyh18@mails., zhao-w19@mails., yezp17@mails., yr16@mails. and liuyongjin@\}tsinghua.edu.cn}}%
\thanks{$^{2}$J. Pan is with the Department of Computer Science, The University of Hong Kong.
        {\tt\small jpan@cs.hku.hk}}%
\thanks{$^{\dag}$Corresponding author}% <-this % stops a space
}

\begin{document}

\maketitle
\thispagestyle{empty}
\pagestyle{empty}

%%%%%%%%%%%%%%%%%%%%%%%%%%%%%%%%%%%%%%%%%%%%%%%%%%%%%%%%%%%%%%%%%%%%%%%%%%%%%%%%
\begin{abstract}
Motion planning for robots of high degrees-of-freedom (DOFs) is an important problem in robotics with sampling-based methods in configuration space $\mathcal{C}$ as one popular solution.
Recently, machine learning methods have been introduced into sampling-based motion planning methods, which train a classifier to distinguish collision free subspace from in-collision subspace in $\mathcal{C}$. In this paper, we propose a novel configuration space decomposition method and show two nice properties resulted from this decomposition. Using these two properties, we build a composite classifier that works compatibly with previous machine learning methods by using them as the elementary classifiers. Experimental results are presented, showing that our composite classifier outperforms state-of-the-art single-classifier methods by a large margin. A real application of motion planning in a multi-robot system in plant phenotyping using three UR5 robotic arms is also presented.
\end{abstract}

%%%%%%%%%%%%%%%%%%%%%%%%%%%%%%%%%%%%%%%%%%%%%%%%%%%%%%%%%%%%%%%%%%%%%%%%%%%%%%%%
\section{INTRODUCTION}
\label{sec:intro}

Motion planning plays an important role in robotics, which finds a collision-free path to move a robot from a source to a target position.
Configuration space $\mathcal{C}$ \cite{Latombe1991} is widely used in robot motion planning, whose spatial dimensions characterize the degrees-of-freedom (DOFs) of the robot and each point in $\mathcal{C}$ represents a configuration of the robot. By decomposing the space $\mathcal{C}$ into a free subspace $\mathcal{C}_{free}$ (i.e., the set of robot configurations without self-collision or collision with obstacles) and an in-collision subspace $\mathcal{C}_{clsn}=\mathcal{C}\setminus\mathcal{C}_{free}$, motion planning is equivalent to finding a path completely within $\mathcal{C}_{free}$.

For robots of high DOFs (e.g., 6-DOF robotic arms or 17-DOF humanoid robots) in complex or dynamic environments, the boundary of $\mathcal{C}_{clsn}$ is very complicated and usually cannot be represented analytically \cite{Latombe1991,TangL04,PanM15}. Sampling-based motion planning (SBMP) methods (e.g., \cite{JailletCS10,KaramanF11ijrr}) were then proposed to use sample points for characterizing $\mathcal{C}_{free}$ and avoid explicitly establishing the boundary of $\mathcal{C}_{clsn}$. Some representative researches include the classic probabilistic roadmaps (PRM) \cite{Kavraki96}, rapidly-exploring random trees (RRT) \cite{Lavalle00,Kuffner00}, the variants of PRM and RRT (e.g., \cite{KaramanF11ijrr}) and some state of the arts \cite{KimKY18icra,Dai19icra}, etc.

To implicitly define $\mathcal{C}_{free}$, usually a large number of samples are required in SBMP methods and each sample is guaranteed to be collision-free by passing an exact collision detector such as GJK \cite{Gilbert98} or FCL \cite{PanCM12icra}, which is very time-consuming. To speed up the collision checking procedure, recently machine learning methods have been introduced into this area; see Section \ref{sec:relate-work} for a summary. These methods use a small subset of samples to train a classifier $F$. Then given an arbitrary sample $s$ in unknown regions in $\mathcal{C}$, the classifier can output a prediction $F(s)$ that serves as a filter to quickly identify obviously in-collision or collision-free samples, and then only a small set of samples with ambiguity need to be finally checked by the exact collision detector. We call these predictions by machine learning methods as {\it approximate} collision checking. To ensure the success of these machine learning methods, the trained classifier must have a high accuracy.

SBMP methods can be used for both single query and multiple queries. For single query, the RRT methods (e.g., \cite{Lavalle00,Kuffner00}) start at a source point in $\mathcal{C}$ and iteratively grow a search tree. At each iteration, a randomly selected point is used to drive the system with a small time step and this leads to a new vertex that is added to the tree by an edge linking it to the nearest vertex in the existing tree. The iteration is terminated when the target point is reached. For multiple queries, the PRM method \cite{Kavraki96} and its variants (e.g., \cite{KaramanF11ijrr}) spread out sample points uniformly covering the whole space $\mathcal{C}_{free}$. Then given any arbitrary source and target points, a collision-free path is obtained by making use of these uniform samples.

The trained classifiers in learning-based methods can improve the efficiency of both single and multiple queries in two ways. The first way is to use the prediction of the classifier to quickly identify the new collision-free samples when adding new vertices into the existing search tree. Pan and Manocha \cite{PanM16ijrr} propose a fast probabilistic collision checking method and prove that the collision query predicted by their classifier converges to the exact collision detection when the size of sampling points increases. Therefore, online learning is important, since the classifier needs to be online updated to improve the classification accuracy when more and more collision-free points are added into the data set. The second way is that instead of using a large number of samples to cover $\mathcal{C}_{free}$, a small set of samples can be used to train the classifier and a large number of samples that are predicted by a classifier as collision-free can be quickly generated by the classifier. After a path is planned using these samples, a final exact collision checking is applied for every sample in the path. For those samples that are in collision, a local repairing operation \cite{arxiv2019} is performed to update the path. In both way, the accuracy and query time of the classifiers are critical to affect the performance of motion planning.

  \begin{figure}[t]
  \centering
  \subfigure[Rendered system design]{\includegraphics[width=.44\columnwidth]{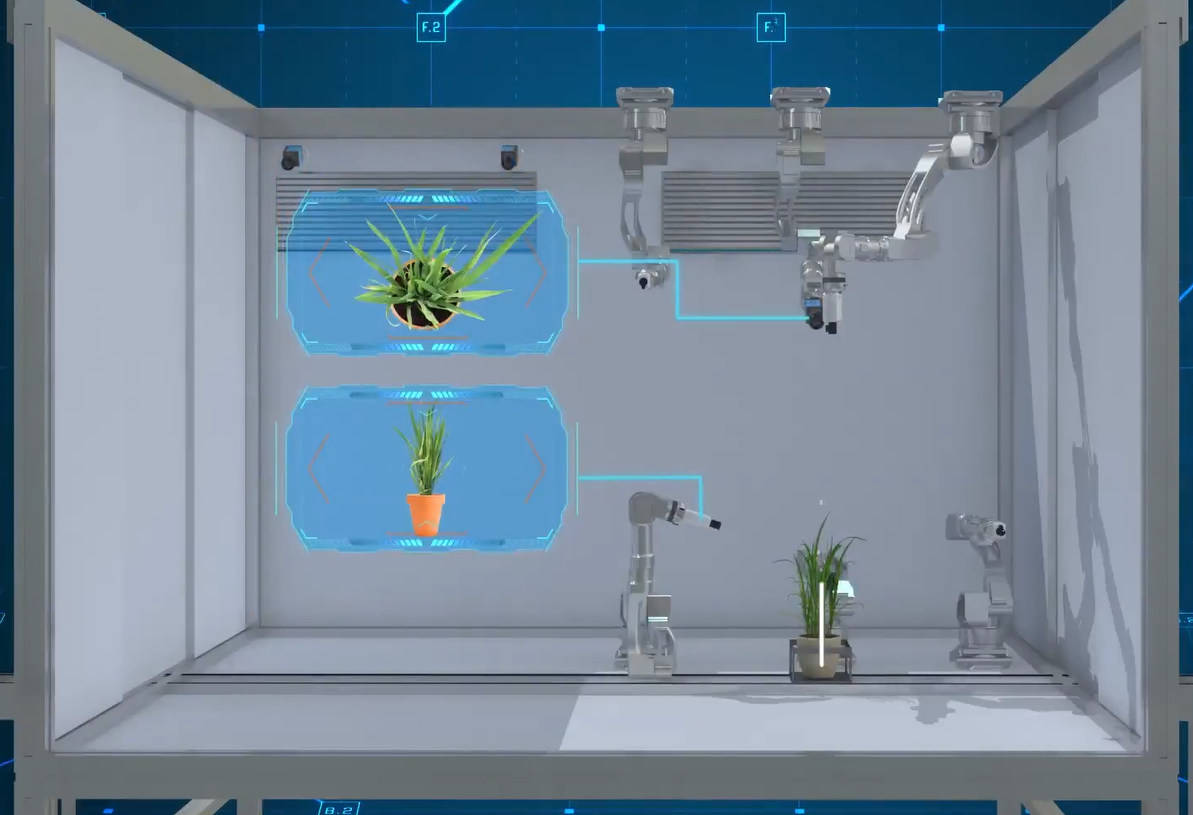}}
  \subfigure[A snapshot of real system]{\includegraphics[width=.54\columnwidth]{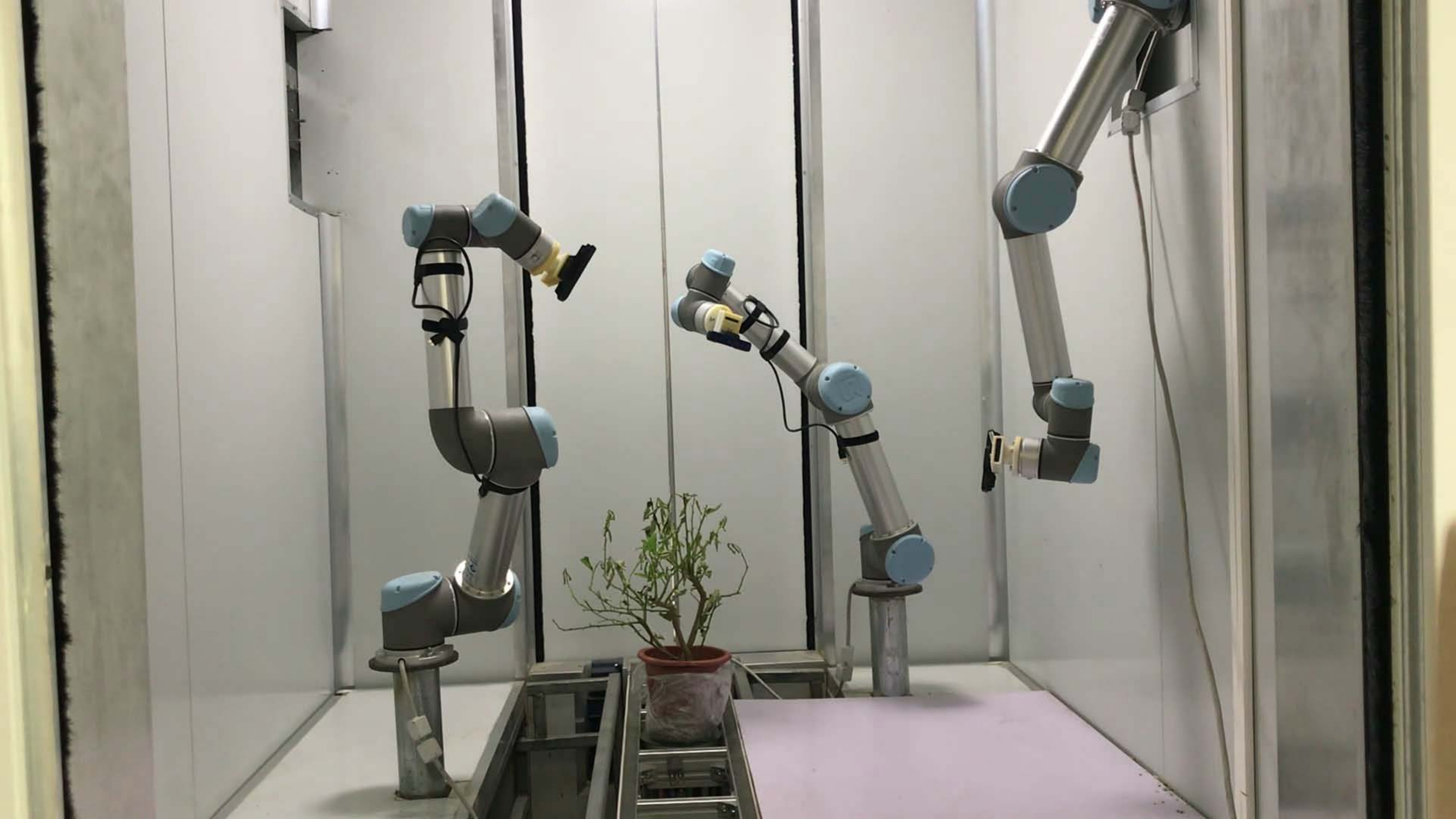}}
  \vspace{-0.15in}
  \caption{An updated multi-robot plant phenotyping system \cite{WuCM19}. By applying our proposed composite classifier, 30\% motion planning time can be saved.
  }
  \label{fig:phenotyping-system}
  \vspace{-0.15in}
  \end{figure}

In most previous machine learning methods, only one binary classifier was trained to predict whether a query sample point $s\in\mathcal{C}$ is in $\mathcal{C}_{free}$ or $\mathcal{C}_{clsn}$. In particular, the subspace $\mathcal{C}_{free}$ or $\mathcal{C}_{clsn}$ is learned as a single entity. In this paper, we propose a novel decomposition of $\mathcal{C}_{free}$ or $\mathcal{C}_{clsn}$, based on the DOFs of the robot. By decomposing $\mathcal{C}_{free}$ or $\mathcal{C}_{clsn}$ into a set of smaller subspaces related to the DOFs of the robot, we construct a composite classifier that consists of a set of simple classifiers and each classifier corresponds to a decomposed subspace. The advantage of this composite classifier is that the set of simple classifiers can be performed as a set of hierarchical filters that quickly filter in-collision samples from easy to hard levels.

Our composite classifier with the configuration-space decomposition scheme is general and can work compatibly with any previous machine learning methods with a single classifier (e.g., \cite{Pan2013,PanM16ijrr,HuhL16icra,arxiv2019}). We show that our composite classifier have a much higher accuracy and averagely 2 time faster than previous single classifier methods. We also apply our composite classifier in the motion planning of an updated multi-robot plant phenotyping system \cite{WuCM19}. This updated system (Figure \ref{fig:phenotyping-system}) consists of three UR5 robotic arms and each arm is equipped with an Intel RealSense SR-300 depth camera. To achieve fast, precise and noninvasive measurements for high-throughput plant phenotyping, all of three arms move simultaneously in each round of phenotyping data acquisition. Our results show that using the proposed composite classifier, 30\% motion planning time can be saved.

\section{RELATED WORK}
\label{sec:relate-work}

Collision detection is frequently called in sampling-based motion planning. In this section, we briefly review the recent machine learning methods that are introduced to speed up the time-consuming collision detection process.

An early work using machine learning is the neural network approach \cite{Garcia2013} that can only handle the collision detection for box-shaped objects.
Pan and Manocha \cite{PanM15} design efficient GPU-based parallel $k$-nearest neighbors (KNN) and parallel collision detection algorithm, and propose an approximation representation for the configuration space based on machine learning. Pan and Manocha \cite{PanM16ijrr} further make use of KNN in online learning configuration space. To achieve fast probabilistic collision checking, they use locality-sensitive hashing techniques that only have a sub-linear time complexity. Their probabilistic collision checking can effectively improve the performance (up to 2x speedup) of various motion planners such as RRT, RRT*, PRM and lazyPRM. Das and Yip \cite{arxiv2019} propose another proxy collision detector that can achieve efficient active learning by utilizing lazy Gram matrix evaluation and a new cheaper kernel to reduce the training and query time.
Heo et al. \cite{Heo19} develop a deep learning method that uses monitoring signals (i.e., external torque at every robotic joint) to estimate collision detection, which is applicable for industrial collaborative robots working with humans.

Most recent researches for speeding up the motion planning in SBMP methods use the idea to train a binary classifier using a small set of samples in $\mathcal{C}$ with correct labels (i.e., in-collision or collision-free) and then approximate collision checking can be quickly obtained by the prediction of the trained classifier.
Various classifiers have been applied, including support vector machine (SVM) \cite{Pan2013}, $k$-nearest neighbor (KNN) \cite{PanM16ijrr}, Gaussian mixture models (GMM) \cite{HuhL16icra,HuhLL17icra} and the Gaussian kernel functions \cite{arxiv2019}. In this paper, we propose a configuration-space decomposition method that leads to a novel composite classifier. This composite classifier consists of a set of binary classifiers and each of them can be any of the above mentioned classifiers, i.e., our model can work compatibly with these previous machine learning methods \cite{Pan2013,PanM16ijrr,HuhL16icra,arxiv2019}. In Section \ref{sec:experiment}, we show that our composite classifier can efficiently reduce the query time and improve the accuracy of single-classifier methods.

\section{DECOMPOSITION OF CONFIGURATION SPACE}

Motion planning in the configuration space $\mathcal{C}$ of high DOFs is a great challenge in robotics. In this paper, we propose a novel decomposition scheme of $\mathcal{C}$ and establish a composite classifier based on the decomposed subspaces, which has significantly better performance than a single classifier directly on $\mathcal{C}$.

\begin{figure}[t]
\centering
%\subfigure[Functionality]{\includegraphics[height=.22\linewidth]{functionality2}}\hspace{2pt}
\includegraphics[width=.56\linewidth]{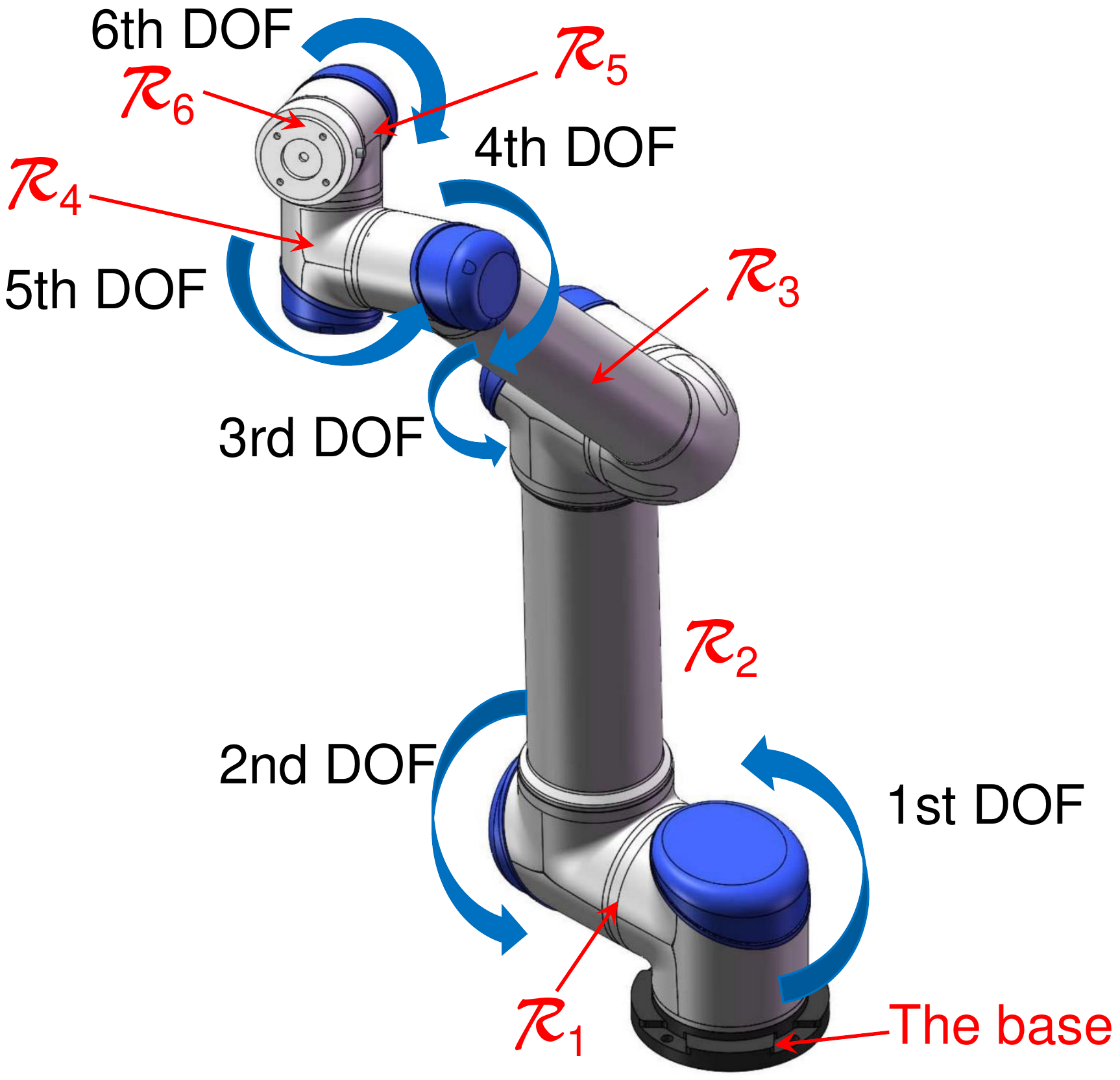}
\vspace{-0.12in}
\caption{The decomposition of a 6-DOF UR5 robot $\mathcal{R}$. We denote these 6 DOFs as $\mathcal{D}=\{d_1,d_2,\cdots,d_6\}$.
The base of UR5 does not move and is not included into $\mathcal{R}$. $\mathcal{R}$ can be decomposed into six components $\mathcal{R}=\cup_{i=1}^6\mathcal{R}_i$.
Each $\mathcal{R}_i$ is associated with a DOF $\mathcal{D}_i=\{d_i\}$, such that the position and orientation of each component $\mathcal{R}_i$ can be uniquely determined by the DOFs $\{\mathcal{D}_1,\mathcal{D}_2,\cdots,\mathcal{D}_i\}$.}
\label{fig:UR5-6DOF}
\vspace{-0.2in}
\end{figure}

Nowadays, robots of high DOFs become ubiquitous, such as robotic arms and humanoid robots. Our work is based on the following important observation. Every robot\footnote{In our study, we only consider the moving part of the robot; e.g., the base of the UR5 in Figure \ref{fig:UR5-6DOF} does not move and is not included.} $\mathcal{R}$ of high DOFs can be separated into disjointed components, satisfying
\begin{equation}
\mathcal{R}=\cup_{k=1}^{n_\mathcal{R}}\mathcal{R}_k \ \mbox{and}\ \mathcal{R}_i\cap\mathcal{R}_j=\emptyset,\ \forall i\neq j,\ i,j\in\{1,2,\cdots,n_\mathcal{R}\}
\label{eq:1}
\end{equation}
where $n_\mathcal{R}$ is the number of components. Let $\mathcal{D}=\{d_1,d_2,\cdots,d_n\}$ be all the DOFs of $\mathcal{R}$, where $n$ is the number of DOFs. For each component $\mathcal{R}_i$, one or more DOFs can be assigned to it, denoted as $\mathcal{D}_i$, such that
\begin{equation}
\mathcal{D}=\cup_{k=1}^{n_\mathcal{R}}\mathcal{D}_k \ \mbox{and}\ \mathcal{D}_i\cap\mathcal{D}_j=\emptyset,\ \forall i\neq j,\ i,j\in\{1,2,\cdots,n_\mathcal{R}\}
\label{eq:2}
\end{equation}
All the components can be ordered in such a way that the position and orientation of each component $\mathcal{R}_i$ can be uniquely determined by the DOFs $\{\mathcal{D}_1,\mathcal{D}_2,\cdots,\mathcal{D}_i\}$. One example of UR5 collaborative robot arm by Universal Robots Corp is shown in Figure \ref{fig:UR5-6DOF}.

The dimension of the configuration space $\mathcal{C}$ of the robot $\mathcal{R}$ is exactly $n$, i.e., the number of DOFs in $\mathcal{R}$.
Most previous works partition $\mathcal{C}$ into a free subspace $\mathcal{C}_{free}$ and an in-collision subspace $\mathcal{C}_{clsn}=\mathcal{C}\setminus\mathcal{C}_{free}$, satisfying that the robot configuration specified by any point in $\mathcal{C}_{free}$ does not have self-collision with $\mathcal{R}$ or collision with obstacles. Based on the characteristics summarized in Eqs. (\ref{eq:1}-\ref{eq:2}), we further decompose $\mathcal{C}_{free}$ and $\mathcal{C}_{clsn}$ accordingly to each DOF of $\mathcal{R}$ as follows.

%For an easy presentation, we consider the static environment, i.e., the obstacles do not move. The dynamic environment will be briefly addressed in xxx.
Throughout this paper, we consider static environment with arbitrary complex obstacles.
First, we consider the DOFs in $\mathcal{D}_1=\{d_1,\cdots,d_{n_1}\}$, where $n_1$ is the number of DOFs in $\mathcal{D}_1$. Let $\mathcal{C}_1$ be the subspace of $\mathcal{C}$ spanned by the DOFs in $\mathcal{D}_1$.
We define the subspace $\mathcal{C}_{1free}$ of $\mathcal{C}_1$ by that the configuration of the component $\mathcal{R}_1$ specified by any point in $\mathcal{C}_{1free}$ does not have self-collision or collision with obstacles. Then we partition $\mathcal{C}_1$ into the free subspace $\mathcal{C}_{1free}$ and the in-collision subspace $\mathcal{C}_{1clsn}=\mathcal{C}_1\setminus\mathcal{C}_{1free}$.
We further define an expansion operation $*$ that expands the dimension of $\mathcal{C}_{1free}$ from $n_1$ to $n$:
\begin{equation}
\begin{array}{l}
\mathcal{C}^*_{1free}=\{\mathcal{C}:\mbox{coordinates in the first}\ n_1\ \mbox{dimensions}\\ \quad \quad \quad \quad \mbox{are restricted in the range of}\ \mathcal{C}_{1free}\}
\end{array}
\end{equation}
Similarly, we define
\begin{equation}
\begin{array}{l}
\mathcal{C}^*_{1clsn}=\{\mathcal{C}:\mbox{coordinates in the first}\ n_1\ \mbox{dimensions}\\ \quad \quad \quad \quad \mbox{are restricted in the range of}\ \mathcal{C}_{1clsn}\}
\end{array}
\label{eq:C*1clsn}
\end{equation}
Obviously, $\mathcal{C}^*_{1clsn}\subseteq\mathcal{C}_{clsn}$.

%Suppose that we have specified $\mathcal{C}_i$, $\mathcal{C}_{ifree}$ and $\mathcal{C}_{iclsn}$, $1\leq i<n$, for all components $\mathcal{R}_1,\cdots,\mathcal{R}_i$. Now we consider the subspace $\mathcal{C}_{i+1}$, whose dimension is the number of DOFs in $\mathcal{D}^{i+1}=\cup_{j=1}^{i+1}\mathcal{D}_j$. Any point in $\mathcal{D}^{i+1}$ specifies the configuration of the component $\mathcal{R}_{i+1}$. We define the subspace $\mathcal{C}_{(i+1)free}$ of $\mathcal{C}_{i+1}$ by that the configuration of the component $\mathcal{R}_{i+1}$ specified by any point in $\mathcal{C}_{(i+1)free}$ does not have self-collision with $\mathcal{R}$ or collision with obstacles.
%Then we partition $\mathcal{C}_{i+1}$ into the free subspace $\mathcal{C}_{(i+1)free}$ and the in-collision subspace $\mathcal{C}_{(i+1)clsn}=\mathcal{C}_{i+1}\setminus\mathcal{C}_{(i+1)free}$.
%Let
%\begin{equation}
%\begin{array}{l}
%\mathcal{C}^*_{(i+1)free}=\{\mathcal{C}:\mbox{coordinates in the first}\ n_{i+1}\\ \mbox{dimensions are restricted in the range of}\ \mathcal{C}^*_{(i+1)free}\}
%\end{array}
%\end{equation}
%\begin{equation}
%\begin{array}{l}
%\mathcal{C}^*_{(i+1)clsn}=\{\mathcal{C}:\mbox{coordinates in the first}\ n_{i+1}\\ \mbox{dimensions are restricted in the range of}\ \mathcal{C}^*_{(i+1)clsn}\}
%\end{array}
%\end{equation}
%where $n_{i+1}$ is the number of DOFs in $\mathcal{D}^{i+1}$.

Now we consider $\mathcal{C}_i$, $1<i\leq n_\mathcal{R}$, which is the subspace spanned by the DOFs in $\mathcal{D}^i=\cup_{j=1}^i\mathcal{D}_j$. We denote the number of DOFs in $\mathcal{D}^i$ as $n_i$. Any point in $\mathcal{D}^i$ specifies the configuration of the component $\mathcal{R}_i$. We define the subspace $\mathcal{C}_{ifree}$ of $\mathcal{C}_i$ by that the configuration of the component $\mathcal{R}_i$ specified by any point in $\mathcal{C}_{ifree}$ does not have self-collision with $\cup_{j=1}^i\mathcal{R}_i$ or collision with obstacles.
Then we partition $\mathcal{C}_i$ into the free subspace $\mathcal{C}_{ifree}$ and the in-collision subspace $\mathcal{C}_{iclsn}=\mathcal{C}_i\setminus\mathcal{C}_{ifree}$.
Let
\begin{equation}
\begin{array}{l}
\mathcal{C}^*_{ifree}=\{\mathcal{C}:\mbox{coordinates in the first}\ n_i\\ \mbox{dimensions are restricted in the range of}\ \mathcal{C}_{ifree}\}
\end{array}
\end{equation}
\begin{equation}
\begin{array}{l}
\mathcal{C}^*_{iclsn}=\{\mathcal{C}:\mbox{coordinates in the first}\ n_i\\ \mbox{dimensions are restricted in the range of}\ \mathcal{C}_{iclsn}\}
\end{array}
\label{eq:C*iclsn}
\end{equation}
where $n_i$ is the number of DOFs in $\mathcal{D}^i$.

\begin{figure*}[t]
\centering
%\subfigure[Functionality]{\includegraphics[height=.22\linewidth]{functionality2}}\hspace{2pt}
\includegraphics[width=.9\linewidth]{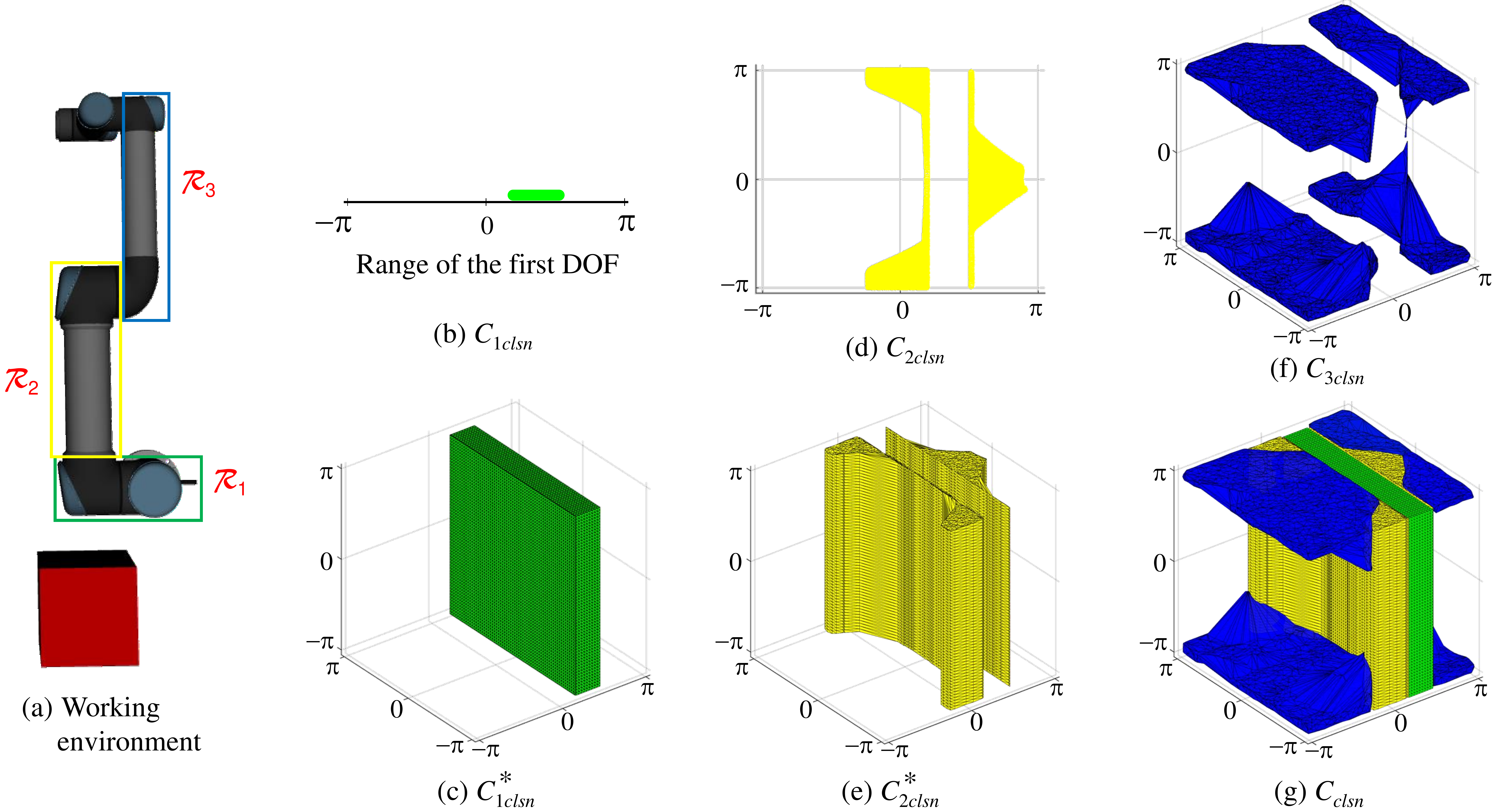}
\vspace{-0.1in}
\caption{(a) For easy illustration, we consider a degenerated 3-DOF UR5 robot $\mathcal{R}=\{\mathcal{R}_1,\mathcal{R}_2,\mathcal{R}_3\}$, with one cubic obstacle (shown in red). Each of three DOFs $\{d_1,d_2,d_3\}$ is ranged from $-\pi$ to $\pi$. Each $\mathcal{R}_i$, $i=1,2,3$, is associated with a DOF $\mathcal{D}_i=\{d_i\}$, such that the position and orientation of $\mathcal{R}_i$ are uniquely determined by the DOFs $\{\mathcal{D}_1,\cdots,\mathcal{D}_i\}$.
(b) The in-collision subspace $\mathcal{C}_{1clsn}$ (shown as the green bold line segment) in $\mathcal{C}_1$ spanned by the DOF $d_1$.
(c) The expanded in-collision subspace $\mathcal{C}^*_{1clsn}$ (shown in green box) in $\mathcal{C}$ as defined in Eq. (\ref{eq:C*1clsn}).
(d) The in-collision subspace $\mathcal{C}_{2clsn}$ (shown as the yellow area) in $\mathcal{C}_2$ spanned by the DOFs $\{d_1,d_2\}$.
(e) The expanded in-collision subspace $\mathcal{C}^*_{2clsn}$ (shown as the yellow volume) in $\mathcal{C}$ as defined in Eq. (\ref{eq:C*iclsn}).
(f) The in-collision subspace $\mathcal{C}_{3clsn}$ (shown as the blue volume) in $\mathcal{C}$ spanned by the DOFs $\{d_1,d_2,d_3\}$.
(g) The final in-collision space $\mathcal{C}_{clsn}=\cup_{i=1}^{3}\mathcal{C}^*_{iclsn}$ (shown as the color volume) in $\mathcal{C}$ as defined in Eq. (\ref{eq:property1}).}
\label{fig:configuration-decomposition}
\vspace{-0.15in}
\end{figure*}

The decomposed subspaces $\{\mathcal{C}^*_{1free},\mathcal{C}^*_{2free},\cdots,\mathcal{C}^*_{n_\mathcal{R}free}\}$ and $\{\mathcal{C}^*_{1clsn},\mathcal{C}^*_{2clsn},\cdots,\mathcal{C}^*_{n_\mathcal{R}clsn}\}$ have the following two important properties:
\begin{equation}
\cup_{i=1}^{n_\mathcal{R}}\mathcal{C}^*_{iclsn}=\mathcal{C}_{clsn}
\label{eq:property1}
\end{equation}
\begin{equation}
\cap_{i=1}^{n_\mathcal{R}}\mathcal{C}^*_{ifree}=\mathcal{C}_{free}
\label{eq:property2}
\end{equation}
See Figure \ref{fig:configuration-decomposition} for an example.

\section{COMPOSITE CLASSIFIER}

In this section, we show that the properties in Eqs. (\ref{eq:property1}-\ref{eq:property2}) can lead to an efficient composite classifier.
Our proposed composite classifier works compatibly with previous machine learning methods that train a single binary classifier to distinguish $\mathcal{C}_{free}$ from $\mathcal{C}_{clsn}$ (e.g., \cite{Pan2013,PanM16ijrr,HuhL16icra,arxiv2019}). Let $f$ be an elementary classifier that can be any one in these previous works.

Given a set of samples with ground-truth labels (in-collision or collision-free) in $\mathcal{C}$, we train an elementary classifier $f_i$ for each robotic component $\mathcal{R}_i$ in the subspace $\mathcal{C}_i$, $i=1,2,\cdots,n_\mathcal{R}$, i.e.,
\begin{equation}
f_i(x)=\left\{
\begin{array}{ll}
1 & \mbox{if}\ x\in\mathcal{C}_{ifree}\\
0 & \mbox{otherwise}
\end{array}
\right.,\ x\in\mathcal{C}_i
\label{eq:ith-element-classifier}
\end{equation}
The our composite classifier $F$ is defined by
\begin{equation}
F=f_1\land f_2\land\cdots\land f_{n_\mathcal{R}}
\end{equation}
where $\land$ is the logistic AND operation, meaning that $F(x)=1$ if and only if all of its operands are true, i.e., $f_i(x)=1$ for $i=1,2,\cdots,n_\mathcal{R}$, $x\in\mathcal{C}$.

We have the following equivalent form of $F$, which can also be implied from the properties in Eqs. (\ref{eq:property1}-\ref{eq:property2}):
\begin{equation}
F=\lnot\left[(\lnot f_1)\lor(\lnot f_2)\lor\cdots\lor(\lnot f_{n_\mathcal{R}})\right]
\end{equation}
where $\lnot$ and $\lor$ are logistic NOT and OR operations.

The advantage of the composite classifier $F$ is of three-fold:
\begin{itemize}
\item Each elementary classifier in $\{f_1,f_2,\cdots,f_{n_\mathcal{R}}\}$ works in a subspace $\mathcal{C}_i$ of $\mathcal{C}$, in which the boundary between $\mathcal{C}_{ifree}$ and $\mathcal{C}_{iclsn}$ is much simpler than the boundary between $\mathcal{C}_{free}$ and $\mathcal{C}_{clsn}$ in $\mathcal{C}$ (see Figures \ref{fig:configuration-decomposition}b, \ref{fig:configuration-decomposition}d, \ref{fig:configuration-decomposition}f, \ref{fig:configuration-decomposition}g for an example), and thus the classification accuracy of each elementary classifier (as well as the composite classifier) is much higher than that of a single classifier directly working on $\mathcal{C}$;
\item Except for $f_{n_\mathcal{R}}$, all other elementary classifiers work in low-dimensional subspaces $\mathcal{C}_i$ with simple class boundaries and thus the classification speed is fast;
\item The elementary classifiers $\{f_1,f_2,\cdots,f_{n_\mathcal{R}}\}$ act as a hierarchical set of filters that filter in-collision samples from the lowest to the highest dimensions, i.e., not all the in-collision samples need to be checked by all the filters. Therefore, the overall classification speed is still faster than a single classifier directly working on $\mathcal{C}$.
\end{itemize}
Our experimental results in Section \ref{sec:experiment} show that averagely the classification accuracy of our composite classifier is improved 10\%-20\% and 2 times faster than a single classifier directly working on $\mathcal{C}$.

To take the full advantage of the proposed composite classifier $F$, it is worthy of noting the following implementation details.

{\it Effective DOFs in $\mathcal{R}$.}
In different application scenarios, not every DOF of the robot $\mathcal{R}$ has the same importance. For example, for the UR5 robot shown in Figure \ref{fig:UR5-6DOF}, if a dexterous hand is attached to the end of component $\mathcal{R}_6$, the rotation of $\mathcal{R}_6$ caused by the 6th DOF may make the dexterous hand collide with obstacle, and thus the 6th DOF should be seriously considered. However, if a cylindric platform (like the one for the 3D printing in \cite{WuDFLW17icra}) is attached to the end of component $\mathcal{R}_6$, the rotation of $\mathcal{R}_6$ caused by the 6th DOF only change the orientation of the cylindric platform, but cannot change its collision status. Therefore, the 6th DOF is not effective. In most cases, it is easy to determine the effectiveness of each DOF in $\mathcal{R}$ using a simple input from the user. For non-effective DOFs $D_j$, we do not build the subspace $C_j$ and train the classifier $f_j$.

{\it Online learning.}
As aforementioned in Section \ref{sec:intro}, to apply the proposed composite classifier $F$ in single query with the RRT methods \cite{Lavalle00,Kuffner00}, it must have the ability of online learning, i.e., efficiently updating $F$ when more and more samples are obtained during the RRT sampling process. Online updating $F$ equals to online updating its elementary classifiers $\{f_1,f_2,\cdots,f_{n_\mathcal{R}}\}$. For commonly used elementary classifiers, their online learning schemes have been proposed, e.g., online learning of SVM \cite{GlasmachersI08neco}, KNN \cite{PanM16ijrr}, GMM \cite{HuhL16icra,HuhLL17icra} and the Gaussian kernel functions \cite{arxiv2019}.

%{\it Dynamic environments.}

\begin{table*}[thb]
\centering
\caption{
The comparison of single-classifier (SC) methods (SVM \cite{Pan2013} and KNN \cite{PanM16ijrr}) and our composite-classifier method using SVM and KNN as elementary classifier respectively in a test environment with 1K, 10K and 100K random samples in the configuration space whose working environment consists of a 6-DOF UR5 robot and a cubic obstacle. The averaged true positive rate (TPR), true negative rate (TNP), classification accuracy, accuracy improvement ($=\frac{\mbox{Our method accuracy}-\mbox{SC accuracy}}{\mbox{SC accuracy}}$), query time (measured by microsecond $\mu s$) and speed up ($\mbox{times}=\frac{\mbox{SC method time}}{\mbox{Our method time}}$) are reported.}
\vspace{-0.1in}
\begin{tabular}{c|c|c|c|c|c|c|c|c|c|c|c}
\hline
Elementary & Number & \multicolumn{2}{c|}{TPR} & \multicolumn{2}{c|}{TNR} & \multicolumn{3}{c|}{Accuracy} & \multicolumn{3}{c}{Query time ($\mu s$)}\\
\cline{3-12}
classifier & of samples & SC & Ours & SC & Ours & SC & Ours & Improvement & SC & Ours & Speed up (times)\\
\hline
{\multirow {3}{*}{SVM}} & 1K & 0.734 & 0.924 & 0.718 & 0.845 & 0.727 & 0.884 & 21.6\% & 36.55 & 19.85 & 1.84x\\
& 10K & 0.812 & 0.956 & 0.800 & 0.940 & 0.807 & 0.949 & 17.6\% & 170.07 & 53.90 & 3.16x\\
& 100K & 0.891 & 0.991 & 0.882 & 0.972 & 0.887 & 0.982 & 10.7\% & 1090.09 & 242.18 & 4.50x\\
\hline
{\multirow {3}{*}{KNN}} & 1K & 0.695 & 0.892 & 0.655 & 0.789 & 0.677 & 0.838 & 23.8\% & 2.63 & 2.34 & 1.12x\\
& 10K & 0.794 & 0.952 & 0.732 & 0.881 & 0.765 & 0.918 & 20.0\% & 6.72 & 5.12 & 1.31x\\
& 100K & 0.850 & 0.971 & 0.828 & 0.952 & 0.840 & 0.962 & 14.5\% & 21.08 & 14.15 & 1.49x\\
\hline
\end{tabular}%
\label{tab:accuracy-time}
\vspace{-0.2in}
\end{table*}

\section{EXPERIMENTS}
\label{sec:experiment}

Using the 6-DOF UR5 robot as the experiment platform, we implement the proposed configuration space decomposition method and the composite classifier $F$ in MATLAB.
To train $F$, we randomly sample the parametric domain of the configuration space with a uniform distribution in a C++ ROS environment,
and the label for each sample is specified by an exact collision detector called Flexible Collision Library (FCL) \cite{PanCM12icra}.
All the running time reported in this section is recoded in a PC with an Intel i7-8700 CPU (3.20GHz) and 16GB RAM.

\subsection{Classification Accuracy and Time}
\label{subsec:classification-accuracy}

We choose two representative single-classifier methods --- SVM \cite{Pan2013} and KNN \cite{PanM16ijrr} --- as the baselines to compare with our composite classifier.
We denote the composite classifier $F$ that uses SVM or KNN as the elementary classifier as $F_{SVM}$ or $F_{KNN}$.
First, we generate a test environment by randomly placing four cubic obstacles around the position of 6-DOF UR5 robot. Then we randomly sample 1K, 10K and 100K points in the configuration space $\mathcal{C}$ as the training set, respectively. To test the classification accuracy, another 10K points are randomly sampled in $\mathcal{C}$.
The averaged classification accuracy and query time are summarized in Table \ref{tab:accuracy-time}, in which the true positive rate (TPR) and the true negative rate (TNP) are also reported.
The following four characteristics are observed from these results.

\begin{table}[thb]
\centering
\caption{In the working environments consisting of a 6-DOF UR5 robot and 1, 2, 4, 8 cubic obstacles, the percentage (\%) of 10K random samples that lie in in-collision and collision-free subspaces is reported. For in-collision samples, the percentage (\%) of them that can be detected in $\mathcal{C}_{1clsn},\mathcal{C}_{2clsn},\cdots,\mathcal{C}_{6clsn}$ is also reported.}
\vspace{-0.08in}
\setlength\tabcolsep{3.5pt}
\begin{tabular}{c|c|c|c|c|c|c|c}
\hline
Obstacle & \multicolumn{6}{c|}{In-collision} & Collision\\
\cline{2-7}
number & $\mathcal{C}_{1clsn}$ & $\mathcal{C}_{2clsn}$ & $\mathcal{C}_{3clsn}$ & $\mathcal{C}_{4clsn}$ & $\mathcal{C}_{5clsn}$ & $\mathcal{C}_{6clsn}$ & free\\
\hline
1 & 16.3 & 14.1 & 9.1 & 1.3 & 7.2 & 0.1 & 51.9\\
2 & 16.0 & 17.9 & 9.3 & 1.6 & 6.4 & 0.2 & 48.7 \\
4 & 16.7 & 19.4 & 9.5 & 2.1 & 5.8 & 0.2 & 46.3 \\
8 & 16.3 & 27.4 & 15.0 & 2.2 & 5.1 & 0.1 & 33.9 \\
\hline
\end{tabular}%
\label{tab:percentage-samples}
\vspace{-0.2in}
\end{table}

First, the accuracy of the composite classifier is much higher (improving 10\%-20\%) than that of the single classifier. This is because that our composite classifier $F$ decomposes the configuration space $\mathcal{C}$ into a set of subspaces $\{\mathcal{C}_1,\mathcal{C}_2,\cdots,\mathcal{C}_{n_\mathcal{R}}\}$, in which each subspace $\mathcal{C}_i$ has a simple boundary between $\mathcal{C}_{ifree}$ and $\mathcal{C}_{iclsn}$, and can be classified much more accurately by an elementary classifier $f_i$, when compared to the accuracy of using a single classifier to directly classify $\mathcal{C}_{free}$ and $\mathcal{C}_{clsn}$ in $\mathcal{C}$.

Second, the query time of the composite classifier is faster than that of the single classifier. This is because that although the composite classifier may need to check multiple elementary classifiers, the query time in each elementary classifier is faster and only a few samples need to pass all these elementary classifiers. To further reveal the latter property, we report the percentage of random samples that lie in in-collision (which can be further decomposed into different subspaces $\{\mathcal{C}_{1clsn},\mathcal{C}_{2clsn},\cdots,\mathcal{C}_{6clsn}\}$) and collision-free subspaces in Table \ref{tab:percentage-samples}, showing that 40\%-60\% samples are filtered by the first three elementary classifiers in the subspaces $\{\mathcal{C}_1,\mathcal{C}_2,\mathcal{C}_3\}$. The results in Table \ref{tab:percentage-samples} also show that the 6th DOF is not an effective DOF and removing the elementary classifier $f_6$ by merging it with $f_5$ can further speed up the query time.

\begin{table}[thb]
\centering
\caption{
The comparison of single-classifier (SC) methods (SVM \cite{Pan2013} and KNN \cite{PanM16ijrr}) and our composite-classifier method using SVM and KNN as elementary classifier respectively in a test environment consisting of a 6-DOF UR5 robot and 2, 4, 8 cubic obstacles. Classification accuracy, accuracy improvement ($\Delta=\frac{\mbox{Our method accuracy}-\mbox{SC accuracy}}{\mbox{SC accuracy}}$), query time (measured by microsecond $\mu s$) and speed up ($\tau=\frac{\mbox{SC method time}}{\mbox{Our method time}}$) are reported.}
\vspace{-0.08in}
\setlength\tabcolsep{4pt}
\begin{tabular}{c|c|c|c|c|c|c|c}
\hline
Elementary & Num of & \multicolumn{3}{c|}{Accuracy} & \multicolumn{3}{c}{Query time ($\mu s$)}\\
\cline{3-8}
classifier & obstacles & SC & Ours & $\Delta$ & SC & Ours & $\tau$\\
\hline
{\multirow {3}{*}{SVM}} & 2 & 0.828 & 0.927 & 12.0\% & 168.75 & 57.25 & 2.95x\\
& 4 & 0.807 & 0.949 & 17.6\% & 170.07 & 53.90 & 3.16x\\
& 8 & 0.768 & 0.908 & 18.2\% & 184.50 & 86.51 & 2.13x\\
\hline
{\multirow {3}{*}{KNN}} & 2 & 0.772 & 0.907 & 17.5\% & 5.12 & 3.46 & 1.48x\\
& 4 & 0.765 & 0.918 & 20.0\% & 6.72 & 5.12 & 1.31x\\
& 8 & 0.733 & 0.896 & 22.2\% & 7.24 & 4.30 & 1.68x\\
\hline
\end{tabular}%
\label{tab:accuracy-time-multiple-obstacle}
\vspace{-0.2in}
\end{table}

Third, the more sampling points, the higher the accuracy of both composite and single classifiers would be. However, even with 100K samples, the accuracy of the composite classifier is still 10\% higher than that of the single classifier.

Fourth, the query time is increased when more samples are used. This is because that the computation of SVM discriminative function involves the weighted average of the kernel functions for all samples and KNN needs to compute the nearest distance to all samples. Our composite classifier is faster than single classifier in terms of both SVM and KNN, but the speed improvement of $F_{SVM}$ on SVM is much better than that of $F_{KNN}$ on KNN. This is because we use a KD-tree to speed up the KNN algorithm but there is no similar data structure for SVM.

\begin{table*}[thb]
\centering
\caption{The success rate (the first number in bracket) and average generating time (the second number in bracket, in millisecond) for motion planning in 100 randomly generated pairs. Each pair consists of two collision-free points (one source and one target) in the configuration space. In the working environments consisting of a 6-DOF UR5 robot and 1, 2, 4, 8 cubic obstacles, respectively. We compare the basic RRT method with FCL, and the variants of RRT methods with learning-based collision checking, including Fastron \cite{arxiv2019}, SVM \cite{Pan2013}, KNN \cite{PanM16ijrr} and our composite classifiers $F_{SVM}$ and $F_{KNN}$.}
\vspace{-0.1in}
\begin{tabular}{c|c|c|c|c|c|c}
\hline
Obstacle & Basic RRT & \multicolumn{5}{c}{Variants of RRT with learning-based collision checking}\\
\cline{3-7}
number & with FCL & Fastron & SVM & $F_{SVM}$ & KNN & $F_{KNN}$\\
\hline
2 & (56\%, 850.5) & (49\%, 524.4) & (38\%, 1329.6) & (73\%, 1058.1) & (37\%, 1330.0) & (51\%, 1130.7)\\
4 & (51\%, 880.9) & (36\%, 562.7) & (35\%, 1592.9) & (69\%, 1101.9) & (39\%, 1866.3) & (46\%, 1216.8)\\
8 & (5\%, 1000.1) & (0\%, not available) & (2\%, 2806.7) & (27\%, 2398.8) & (6\%, 2090.1) & (14\%, 1529.0)\\
\hline
\end{tabular}%
\label{tab:motion-planning}
\vspace{-0.12in}
\end{table*}

To further test our proposed composite classifier in diverse environments, we randomly place 2, 4 and 8 cubic obstacles around the position of 6-DOF UR5 robot.
we randomly sample 10K points in the configuration space $\mathcal C$ as the training set and randomly sample another 10K points for testing.
The accuracy and query time are summarized in Table \ref{tab:accuracy-time-multiple-obstacle}, from which the same conclusion can be drawn as those from Table \ref{tab:accuracy-time}.

\subsection{Motion Planning Efficiency}

As aforementioned in Section \ref{sec:intro}, the learning-based collision checking can help improve the efficiency of both single-query and multiple-query sampling-based motion planning. In this section, we use the single-query RRT method as the baseline for comparison.

We use the open motion planning library (OMPL\footnote{http://ompl.kavrakilab.org/}) \cite{Sucan12}, which provides an optimized implementation of the RRT method. The RRT algorithm uses a biased search to quickly explore the large unsearched space. Using enough time, RRT will eventually build a random space-filling tree and thus find a path between source and target points. For practical usage, we set the parameter {\it MaxTime} (i.e., maximum planning time used by RRT) to be 3 seconds. Therefore, the faster collision detection, the higher the success rate of path planning.

  \begin{figure}[t]
  \centering
  \includegraphics[width=.48\columnwidth]{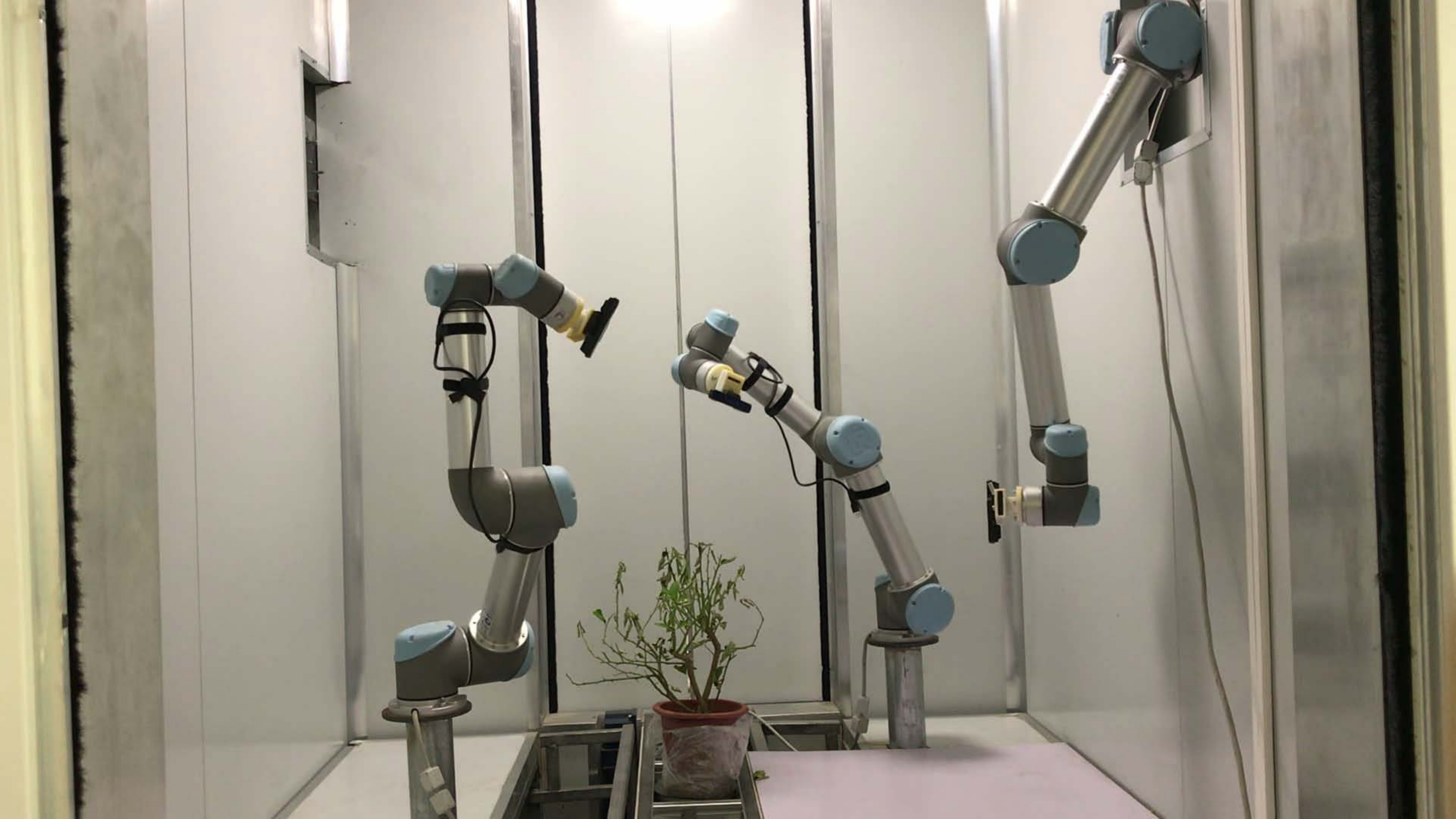}\hspace{6pt}\includegraphics[width=.48\columnwidth]{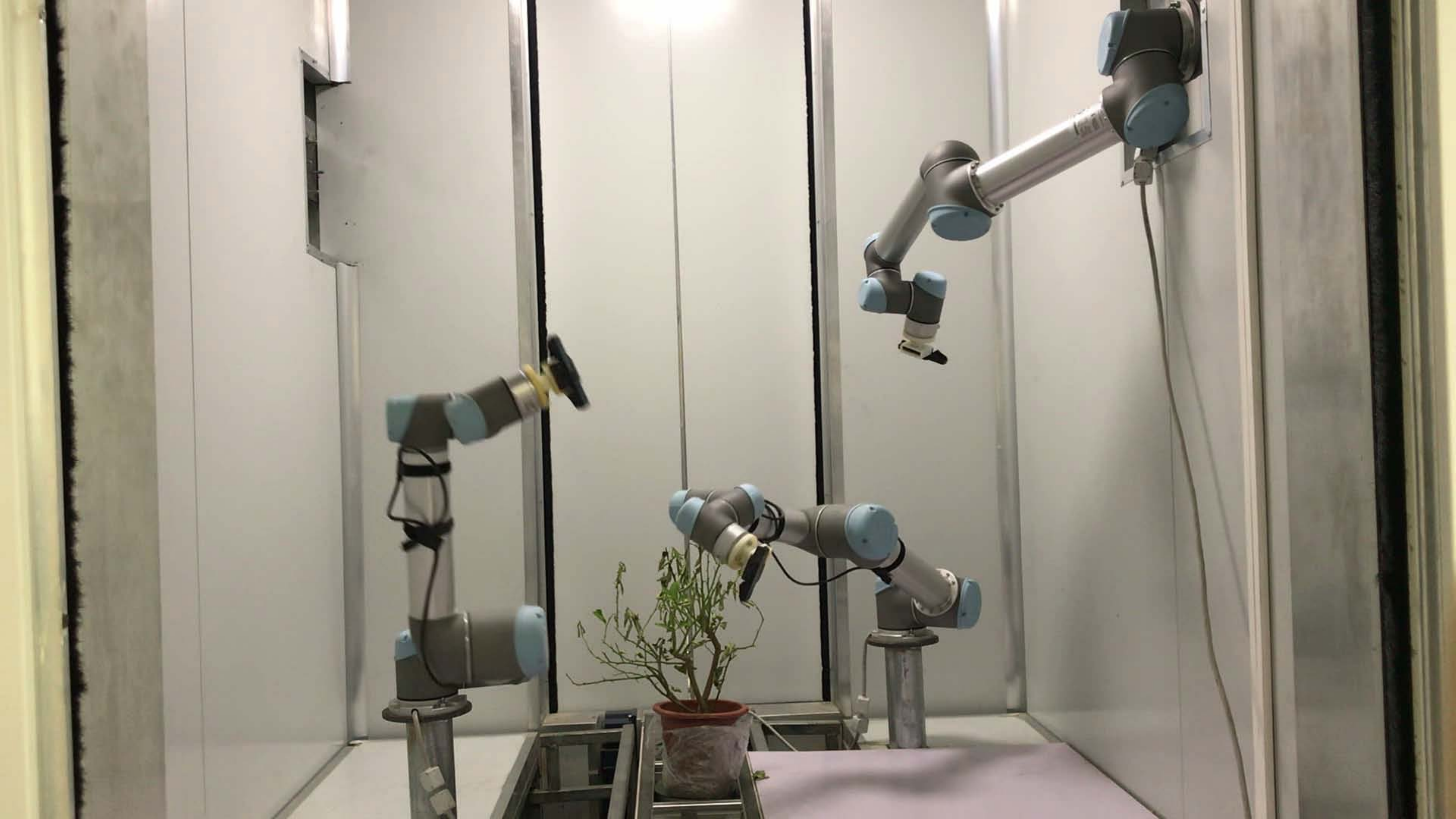}\vspace{6pt}
  \includegraphics[width=.48\columnwidth]{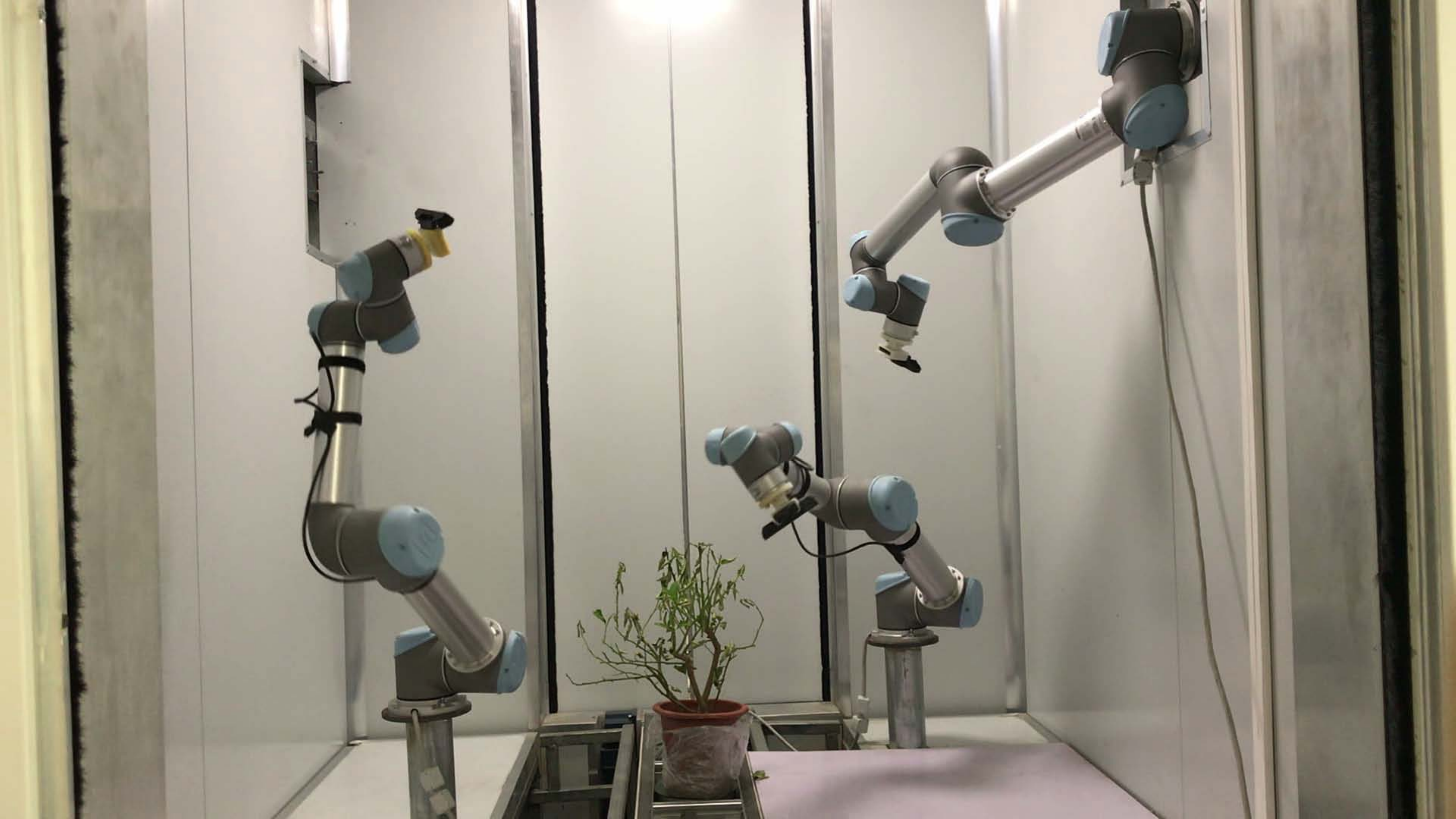}\hspace{6pt}\includegraphics[width=.48\columnwidth]{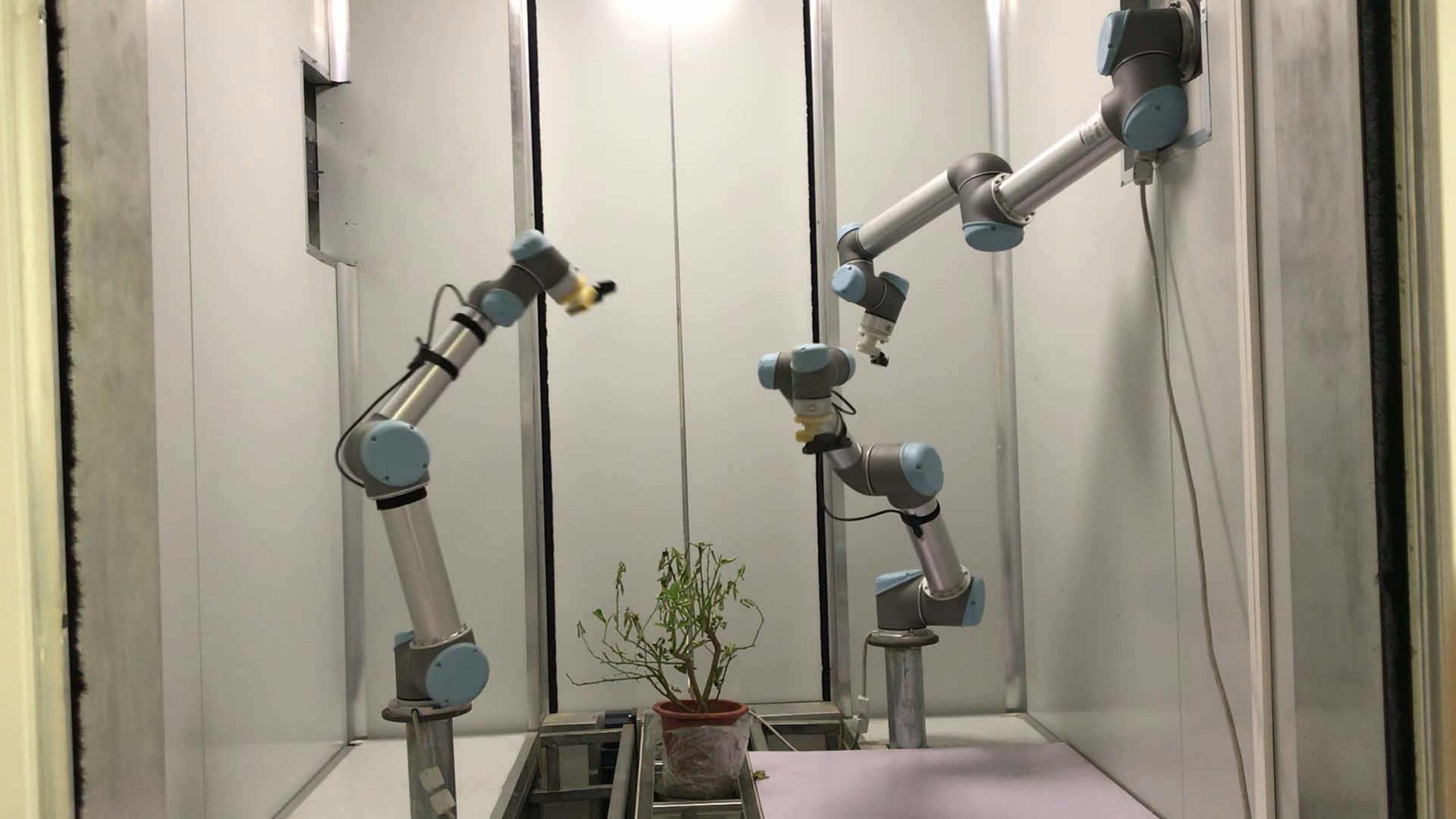}
  \vspace{-0.15in}
  \caption{Snapshots of a motion sequence of an updated multi-robot plant phenotyping system \cite{WuCM19} using the learning-based method with our composite classifier. See accompanying demo video for more details.}
  \label{fig:system-motion}
  \vspace{-0.2in}
  \end{figure}

We set up the basic RRT method implemented by OMPL for single-query motion planning by using FCL \cite{PanCM12icra} for exact collision detection, which is very time-consuming. For comparison, we replace FCL by four learning-based collision detectors, i.e., Fastron \cite{arxiv2019}, SVM \cite{Pan2013}, KNN \cite{PanM16ijrr} and our composite classifiers $F_{SVM}$ (using SVM as the elementary classifier) and $F_{KNN}$ (using KNN as the elementary classifier). Note that the prediction by trained classifiers only provides approximate collision checking. Once a path is planned by RRT with approximate collision checking, a final exact collision checking by FCL is needed for every sample in the path. For those samples that are in-collision, a local repairing operation proposed in the implementation of \cite{arxiv2019} is evoked. Therefore, the more accurate the classifier, the fewer samples need to be repaired and the more efficient the motion planning with the learning-based method would be.

As indicated in Section \ref{subsec:classification-accuracy}, our proposed composite classifier has much better accuracy and less query time than each single classifier, and then will lead to a more efficient motion planning process. This conclusion is demonstrated by the motion planning results summarized in Table \ref{tab:motion-planning}. These results are generated in the same working environments consisting of a 6-DOF UR5 robot and 1, 2, 4, 8 cubic obstacles as in Section \ref{subsec:classification-accuracy}. We randomly generate 100 pairs of collision-free source and target points in the configuration space. The motion planning using different classifiers is applied to plan a path between each pair of source and target points. In the limited maximum planning time (3 seconds), not every pair can have a successful motion planning and we define the {\it success rate} as the ratio $\frac{s}{100}$, where $s$ is the number of pairs that have a successful motion planning and 100 is the number of total pairs. We also define the {\it average generating time} as the time of generating successful paths averaged on $s$ successful paths. The results in Table \ref{tab:motion-planning} show that (1) the larger the number of obstacles, the lower the success rate, (2) our composite classifier can significantly improve the success rate of single classifiers, (3) RRT with composite classifier $F_{SVM}$ significantly improves the success rate of the basic RRT with FCL, and (4) the average generating time of RRT with composite classifier is much shorter than that of RRT with each individual classifier.

\subsection{Application}

We apply the proposed composite classifier in an updated multi-robot plant phenotyping system \cite{WuCM19} for improving the efficiency of motion planning. This system was designed to provide fast, precise and noninvasive measurements for robot-assisted high-throughput plant phenotyping. To achieve this goal, this system was equipped with three UR5 robotic arms that can be moved simultaneously in each round of phenotyping data acquisition, using the depth camera (Intel RealSense SR-300) mounted at each robotic arm. The configuration space of this system has a very complicated boundary between $\mathcal{C}_{free}$ and $\mathcal{C}_{clsn}$ due to high possibilities of collision (with plant leaves) and self-collision (among robotic arms); see Figure \ref{fig:system-motion} for a motion planning example. To satisfy the requirement of high-throughput plant phenotyping, the time for motion planning has to be fast. Using state-of-the-art RRT implementation in OMPL, the average motion planning time for each round is about one second, while using the learning-based method with our composite classifier, the average motion planning time is shorten to 0.692 seconds.

\section{CONCLUSIONS}

In this paper, we propose a simple yet effective configuration space decomposition method based on the chacteristics inherent in the DOFs of robot, which leads to an efficient composite classifier with a much better classification accuracy than previous single classifiers. Our method can work compatibly with previous method by using them as elementary classifiers. Experimental results on both artificial environments with increasing complexity and a real environment of an updated multi-robot plant phenotyping system \cite{WuCM19} demonstrate that our method can effectively shorten the time of motion planning by a large margin.

\addtolength{\textheight}{-10.5cm}   % This command serves to balance the column lengths

%%%%%%%%%%%%%%%%%%%%%%%%%%%%%%%%%%%%%%%%%%%%%%%%%%%%%%%%%%%%%%%%%%%%%%%%%%%%%%%%

%%%%%%%%%%%%%%%%%%%%%%%%%%%%%%%%%%%%%%%%%%%%%%%%%%%%%%%%%%%%%%%%%%%%%%%%%%%%%%%%

%%%%%%%%%%%%%%%%%%%%%%%%%%%%%%%%%%%%%%%%%%%%%%%%%%%%%%%%%%%%%%%%%%%%%%%%%%%%%%%%
%\section*{APPENDIX}


%\section*{ACKNOWLEDGMENT}

%The preferred spelling of the word.

\bibliographystyle{IEEEtran}
\bibliography{rootbib}

\end{document}